\pdfoutput=1

\documentclass[11pt]{article}

\usepackage{ACL2023}

\usepackage{times}
\usepackage{latexsym}
\usepackage{fontawesome}
\usepackage{amsmath}
\usepackage{bm}
\usepackage{amsfonts}
\usepackage{amssymb}
\usepackage{booktabs}
\usepackage{subcaption}
\usepackage{caption}
\usepackage{graphicx}
\usepackage{multirow}
\usepackage[T1]{fontenc}

\usepackage[utf8]{inputenc}

\usepackage{microtype}

\usepackage{inconsolata}
\usepackage{makecell}
\usepackage{xcolor}
\usepackage{todonotes}
\newcommand\blfootnote[1]{%
  \begingroup
  \renewcommand\thefootnote{}\footnote{#1}%
  \addtocounter{footnote}{-1}%
  \endgroup
}

\title{Recall, Expand and Multi-Candidate Cross-Encode: \\
Fast and Accurate Ultra-Fine Entity Typing}

\author{
    \textbf{Chengyue Jiang$^\Diamond$$^\ddagger$},
    \textbf{Wenyang Hui$^\Diamond$$^\ddagger$},
    \textbf{Yong Jiang},
    \textbf{Xiaobin Wang},
    \textbf{Pengjun Xie},
    \textbf{Kewei Tu$^\ddagger$}\thanks{$~~$ Kewei Tu is the corresponding author.} \\
    $^\ddagger$ School of Information Science and Technology, ShanghaiTech University \\
    Shanghai Engineering Research Center of Intelligent Vision and Imaging \\
    \texttt{\{jiangchy,huiwy,tukw\}@shanghaitech,edu.cn;} \\
    \texttt{\{jiangyong.ml,xpjandy\}@gmail.com;} \\
    \texttt{czwangxiaobin@foxmail.com;}
}

\begin{document}
\maketitle

\blfootnote{$^\Diamond$ Equal Contribution.}
\newcommand{\code}{\url{http://github.com/modelscope/AdaSeq/tree/master/examples/MCCE}}
\newcommand{\name}{{MCCE}}

\begin{abstract}
Ultra-fine entity typing (UFET) predicts extremely free-formed types (e.g., {\it president, politician}) of a given entity mention (e.g., {\it Joe Biden}) in context. State-of-the-art (SOTA) methods use the cross-encoder (CE) based architecture. CE concatenates the mention (and its context) with each type and feeds the pairs into a pretrained language model (PLM) to score their relevance. It brings deeper interaction between mention and types to reach better performance but has to perform $N$ (type set size) forward passes to infer types of a single mention. CE is therefore very slow in inference when the type set is large (e.g., $N=10k$ for UFET). 
To this end, we propose to perform entity typing in a recall-expand-filter manner. The recall and expand stages prune the large type set and generate $K$ ($K$ is typically less than $256$) most relevant type candidates for each mention. At the filter stage, we use a novel model called {\name} to concurrently encode and score these $K$ candidates in only one forward pass to obtain the final type prediction. 
We investigate different variants of \name\  and extensive experiments show that \name\  under our paradigm reaches SOTA performance on ultra-fine entity typing and is thousands of times faster than the cross-encoder. We also found \name\ is very effective in fine-grained (130 types) and coarse-grained (9 types) entity typing. Our code is available at \code.
\end{abstract}
\section{Introduction}
Ultra-fine entity typing (UFET) \cite{ufet} aims to predict extremely fine-grained types ({\it e.g., president, politician}) of a given entity mention within its context. It provides detailed semantic understandings of entity mention and is a fundamental step in fine-grained named entity recognition \cite{fget}, and can be utilized to assist various downstream tasks such as relation extraction \cite{fewrel}, keyword extraction \cite{huang2020ner} and content recommendation \cite{upadhyay2021explainable}.

\begin{figure}
    \centering
    \scalebox{0.3}{
    \includegraphics{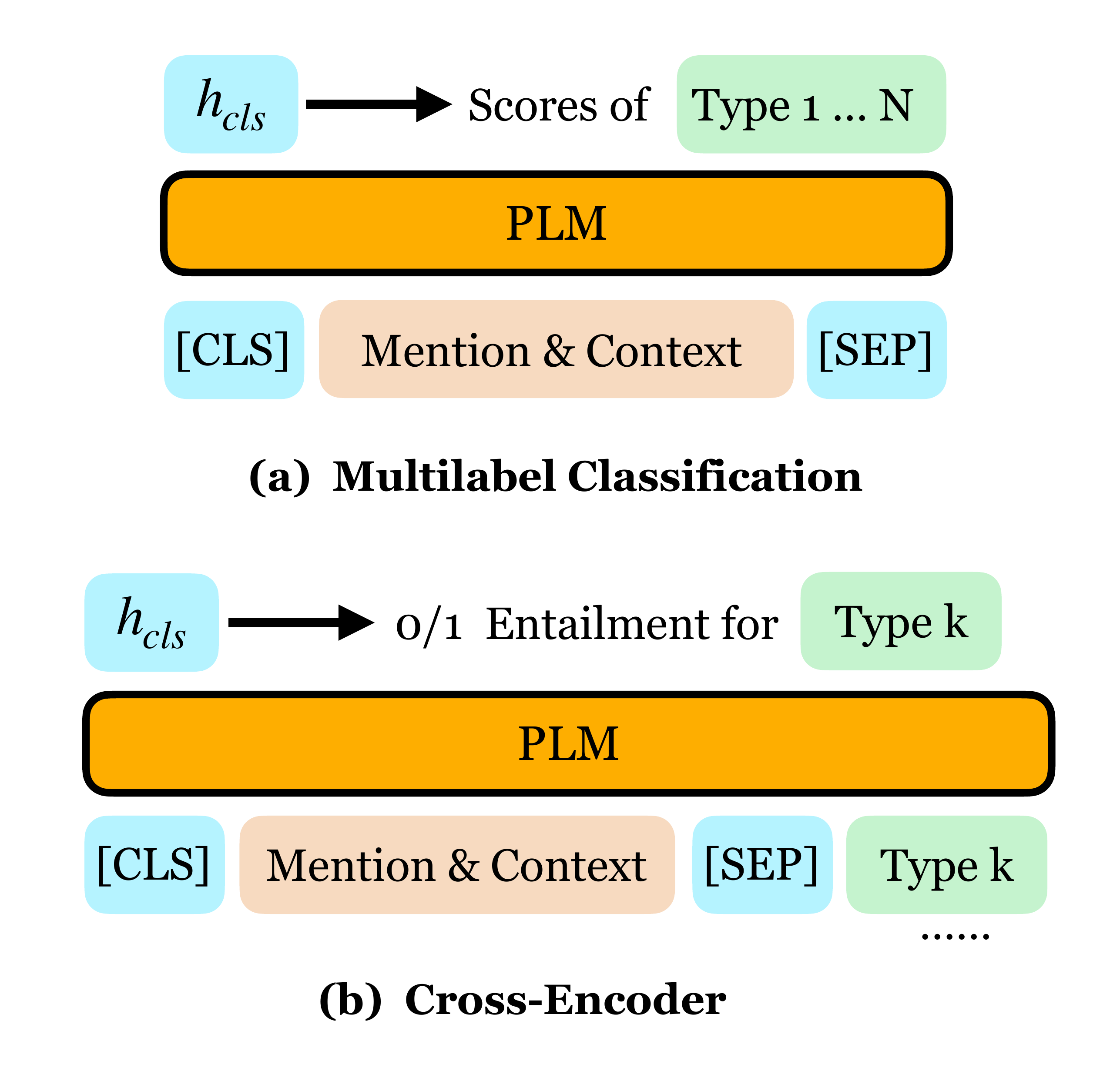}}
    \caption{Cross-Encoder and multi-label classification.}
    \label{fig:mlc_ce}
\end{figure}

\begin{figure*}[t]
    \centering
    \scalebox{0.28}{
    \includegraphics{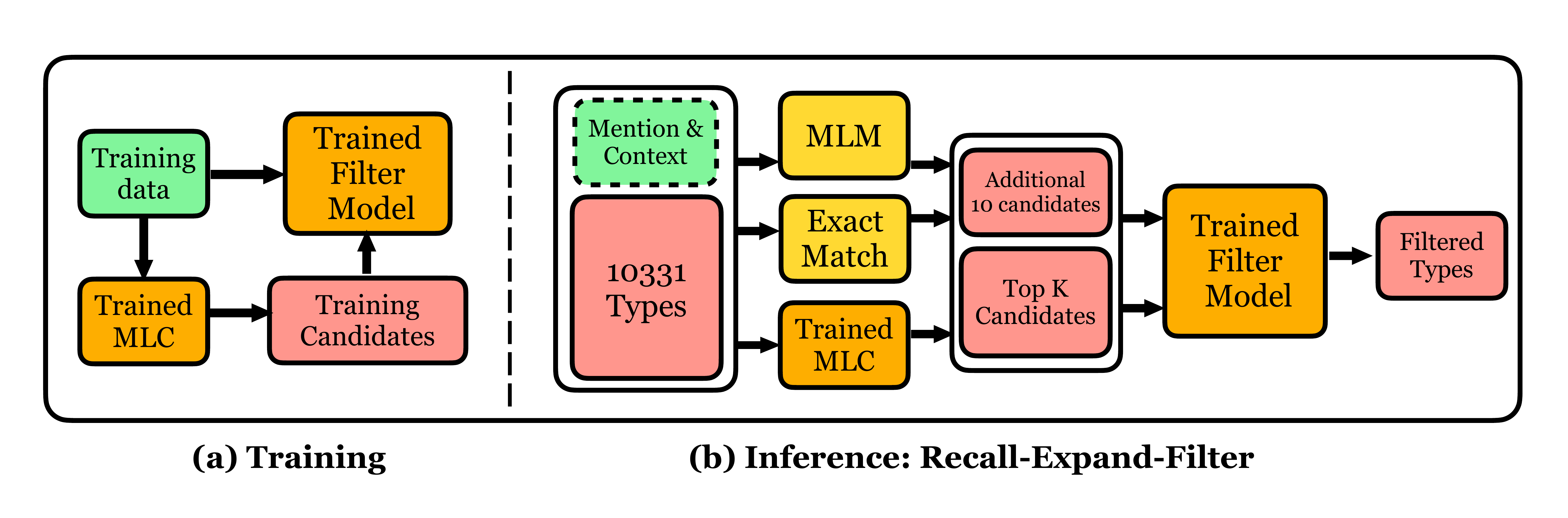}}
    \caption{Training and inference of the recall-expand-filter pradigm.}
    \label{fig:paradigm}
\end{figure*}

Most recently, the cross-encoder (CE) based method \cite{lite} achieves the SOTA performance in UFET. Specifically, \citet{lite} proposed to treat the mention with its context as a premise, and each ultra-fine-grained type as a hypothesis. They then concatenate them together as input and feed it into a pretrained language model (PLM) (e.g., RoBERTa \cite{liu2019roberta}) to score the entailment of mention-type pair as illustrated in Figure \ref{fig:mlc_ce}(b). Compared to the traditional multi-label classification method (shown in Figure \ref{fig:mlc_ce}(a)) that simultaneously scores all types using the mention representation, CE incorporates type semantics in the inference process and enables deeper interactions between types and mention to achieve better performance. However, the CE architecture is slow in inference because it has to enumerate all types (up to 10$k$ types) and score entailment of them given the mention as a premise. There is also no direct interaction between types in CE and is therefore unable to model correlations between types (e.g., one has to be a person if he or she is categorized as a politician), which has been proved to be useful in previous works \cite{npcrf, xiong-etal-2019-imposing}.

\begin{figure}[t]
    \centering
    \scalebox{0.22}{
    \includegraphics{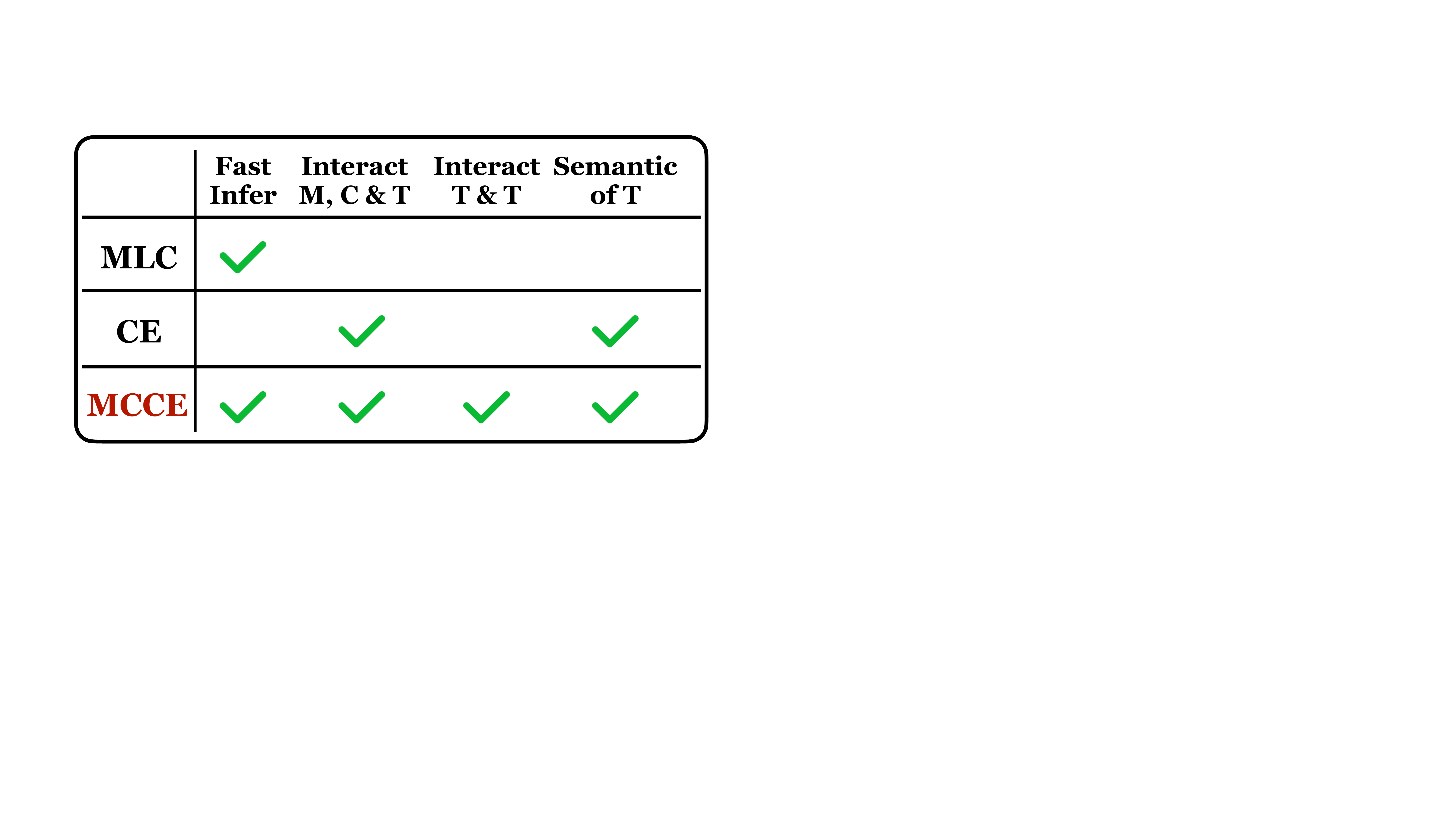}}
    \caption{Comparison of different models, M, C, and T are abbreviations of mention, context, and type.}
    \label{fig:adv}
\end{figure}

To this end, we propose a recall-expand-filter Paradigm for UFET (illustrated in Figure \ref{fig:paradigm}) and a novel model called {\bf \textsc{\name}} for faster and more accurate ultra-fine entity typing. As the name suggests, we first train a multi-label classification (MLC) model to efficiently \textbf{recall} top $K$ candidate types which reduce the number of potential types from thousands to hundreds. As the MLC model recalls candidates based on representations learned from the training data, it's hard to recall candidates that are scarce or unseen in the training set. To this end, we apply a multi-way type candidate \textbf{expansion} step utilizing lexical information and weak supervision from masked language models \cite{mlmet} to improve the recall rate of the candidate set. Last but not least, we propose a backbone called multi-candidate cross-encoder ({\bf \textsc{\name}}) to concurrently encode and \textbf{filter} the expanded type candidate set. Different from CE,  ({\bf \textsc{\name}}) concatenates all recalled type candidates to the mention and its context. The concatenated input is then fed into a PLM to obtain candidate representations and candidate scores. The {\bf \textsc{\name}} architecture allows us to infer types simultaneously from the candidate set while preserving the advantages of CE. Concatenating all candidates also enables {\bf \textsc{\name}} implicitly learns the correlation between types. The advantages of {\bf \textsc{\name}} over existing architectures are shown in Figure \ref{fig:adv}. We also comprehensively investigate the performance and efficiency of {\bf \textsc{\name}} with different input formats and attention mechanisms.

Experiments on two UFET datasets show that {\bf \textsc{\name}} and its variants under our recall-expand-filter paradigm reach SOTA performance and are thousands of times faster than the CE-based previous SOTA method. We also found {\bf \textsc{\name}} is still effective in fine-grained (130 types) and coarse-grained (9 types) entity typing. Our code is available at \code.

\section{Background}
\subsection{Problem Definition}
Given an entity mention $m_i$ within its context sentence $c_i$, ultra-fine entity typing (UFET) aims to predict its correct types $y^g_i \subset \mathcal{Y}$, where $y_i^g$ is the gold types of the $i$-th mention and is a subset of a large type set $\mathcal{Y}$ ($|\mathcal{Y}|$ can be larger than 10$k$). As $|y_i| > 1$ in most cases, UFET can be categorized as a multi-label classification problem. We show statistics of two UFET datasets: {\bf \textsc{UFET}} \cite{ufet} and {\bf \textsc{CFET}}\footnote{As there is no official split available for {\bf \textsc{CFET}}, we split it by ourselves and will release our split in our code.} \cite{cfet} in Table \ref{tab:stat}.

\begin{table}[t]
\scalebox{0.9}{
\centering
\begin{tabular}{ccccc} 
\toprule
dataset & $\vert \mathcal{Y} \vert$      & $\text{avg}(\vert y_i^g \vert)$ & train/dev/test & Lang \\ \midrule
{\bf \textsc{UFET}}   & 10331       & 5.4   & 2k/2k/2k     & EN      \\
{\bf \textsc{CFET}}  & 1299          & 3.5   &   3k/1k/1k & ZH            \\ \bottomrule
\end{tabular}}
\caption{$\text{avg}(\vert y_i^g \vert)$ denotes the average number of gold types per instance, ZH for Chinese.}
\label{tab:stat}
\end{table}

\subsection{Multi-label Classification Model for UFET}
\label{sec:mlc}
Multi-label classification models are widely adopted as backbones for UFET \cite{ufet, onoe-durrett-2019-learning, box4types}. They use an encoder to obtain the mention representation and use a decoder (e.g., MLP) to score types simultaneously. Figure \ref{fig:mlc_ce}(a) shows a representative multi-label classification model adopted by recent methods \cite{npcrf,mlmet}. The contextualized mention representation is obtained by feeding $c_i$ and $m_i$ into the pretrained language models (PLM), and taking the last hidden state of {\tt [CLS]}, $h_{cls}$. The mention representation is then fed into an MLP layer to concurrently obtain all type scores $s_1, \cdots s_N, N=|\mathcal{Y}|$. We call this model MLC and describe its inference and training below. 
\paragraph{MLC Inference} For inference, types with probability higher than a threshold $\tau$ are predicted: $\mathcal{Y}_i = \{y_j | \sigma(s_j) > \tau \}$, $\sigma$ is the sigmoid function. The threshold is tuned on the development set.
\paragraph{MLC Training} Binary Cross-Entropy (BCE) loss between the predicted scores and the gold types are used to train the MLC model: $\mathcal{L}_i = -\frac{1}{N} \sum_{j=1}^N  \alpha \cdot I_{j} \log \sigma(s_j) + (1-I_{j}) \log (1-\sigma(s_j)) $, where $I_j$ is the indicator of $y_j$ being one of the gold types ($y_j \in y_i^g$), and $\alpha$ is a hyper-parameter balancing the loss of positive and negative types. MLC is very efficient in inference. However, the interactions between mention and types in MLC are weak, and the correlations between types are ignored \cite{box4types,xiong-etal-2019-imposing,npcrf}. MLC also has difficulty in integrating type semantics \cite{lite}.

\subsection{Vanilla Cross-Encoders for UFET}
\label{sec:vanilla_ce}
\citet{lite} first proposed to use Cross-Encoder (CE) for UFET. As shown in Figure \ref{fig:mlc_ce}(b), CE concatenates $m_i, c_i, y_j$ together and feeds them into a PLM to obtain the {\tt [CLS]} embedding, then an MLP layer is used to obtain the score of $y_j$ given $m_i, c_i$.
\begin{align}
        \bm{h}_{cls, i} &= \texttt{PLM}(\texttt{[CLS]} \ c_i \ \texttt{[SEP]}\  m_i \ \texttt{[SEP]} \ y_j\ ) \\
        s_j &= \texttt{MLP}(\bm{h}_{cls, i})
\end{align}
The concatenation allows deeper interaction between mention, context, and types (modeled by the multi-head self-attention in PLMs), and also incorporates type semantics.
\paragraph{CE Inference} CE predicts types of a single input $(m_i, c_i)$ by concatenating the input with all possible types $y_j \in \mathcal{Y}$ one by one to predict the scores $s_1, \cdots, s_j$ for each type. Similar to MLC, types that have a higher probability than a threshold are predicted $\mathcal{Y}_i = \{y_j | \sigma(s_j) > \tau \}$. CE requires $N$ forward passes to infer types of a single mention, its inference speed is very slow when $N$ is large.
\paragraph{CE Training} CE is typically trained with marginal ranking loss \cite{lite}. A positive type $y_+ \in y^g_i$ and a negative type $y_{-} \not \in y^g_i$ are sampled from $\mathcal{Y}$ for each data point $(m_i, c_i)$. The loss is computed as:
$$ L_i = \max(\sigma(s_{-}) - \sigma(s_{+}) + \delta, 0) $$ where $s_{+}, s_{-}$ are scores of the sampled positive and negative types, and $\delta$ is the margin tuned on the development set determine how the positive and negative samples should be separated.
\section{Methodology}
\begin{figure*}[t]
    \centering
    \scalebox{0.28}{
    \includegraphics{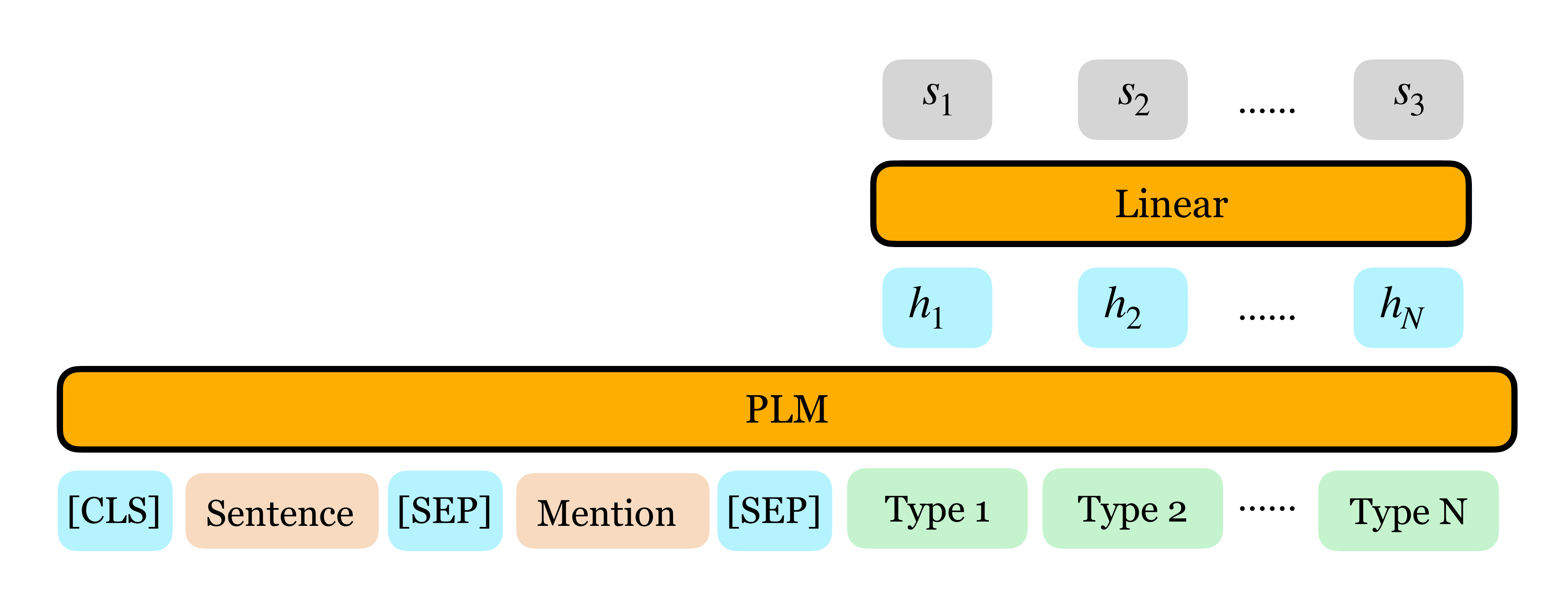}}
    \caption{Multi-candidate cross-encoder (MCCE).}
    \label{fig:ccf}
\end{figure*}
Inspired by techniques in information retrieval \cite{ir} and entity linking \cite{wu2019zero}, we decompose the training and inference of UFET into three stages as illustrated in Figure \ref{fig:paradigm}: (1) Recall stage to reduce the type candidate size (e.g., from $N=10k$ to $K=100$) while guaranteeing the recall rate by an efficient MLC model. (2) Expand stage to incorporate lexical information using exact matching and weak supervision \cite{mlmet} from large pretrained language models such as BERT-Large \cite{bert} to improve recall rate. (3) Filter stage to filter the expanded type candidates to obtain final prediction. For the filter stage, we propose an efficient model: Multi-Candidate Cross-Encoder (MCCE) to concurrently encode and filter type candidates of a given mention with only a single forward pass. 
\subsection{Recall Stage}
To prune the type candidates set, we train a very efficient MLC model introduced in Sec. \ref{sec:mlc} and select the model based on the recall rate (e.g. recall@64) on the development set. Then we use it to infer the top $K_1$ (typically less than 256) candidates $\mathcal{C}_i^R$ for each data point $(m_i, c_i)$ for train, development, and test set. We compare MLC with a widely-used baseline model BM25 \cite{bm25} and show its advantages in Sec. \ref{sec:recall}. 
\subsection{Expand Stage}
\begin{figure}[h]
    \centering
    \scalebox{0.2}{
    \includegraphics{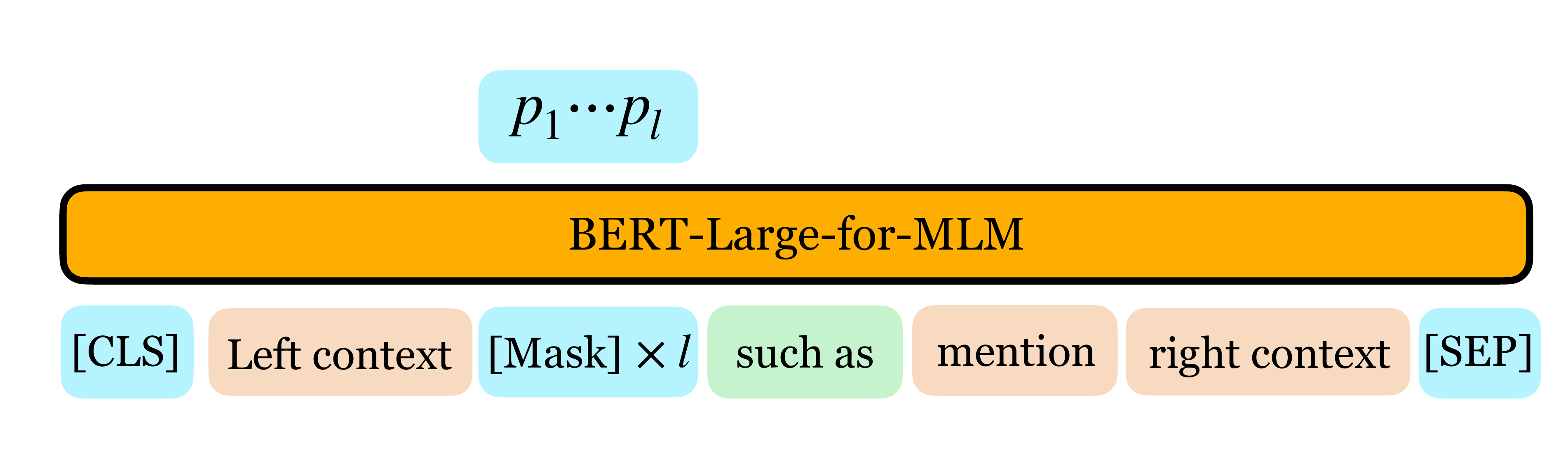}}
    \caption{Recall from MLM using prompts.}
    \label{fig:prompt_recall}
\end{figure}

Due to the lack of training data per type, we found that the MLC we used in the recall stage easily overfits the train set, and is hard to predict the types that only appear in the development and test set. In {\bf \textsc{UFET}} dataset, 30\% of the types in the development set are unseen. To this end, we utilize lexical information using exact match and weak supervision from the masked language model (MLM) to expand the recalled candidates. Both exact match and MLM are able to recall unseen type candidates without any training. 
\paragraph{Exact Match} MLC and Bi-Encoder recall candidates by dense representations. They are known weak at identifying and utilizing the lexical matching information between the input and types \cite{matching_info1, matching_info2}. However, 
 types are free-formed in UFET (e.g., \textit{president, businessman}), and are very likely to appear in the context or mention (e.g., the mention is \textit{`the \textbf{president} Joe Biden'}). To this end, we first find all nouns in the context and mention by NLTK\footnote{nltk.tag package \url{https://www.nltk.org}} POS tagger and normalize their forms, then we recall types that exactly matched with these nouns.
\paragraph{Weak Supervision from MLM} Inspired by recently prompt-based methods for entity typing \cite{ding2021prompt, dfet}, we recall candidates by asking PLMs to fill masks in prompts. Suppose a type $y_j \in \mathcal{Y}$ can be tokenized into $l$ subwords $w_1, \cdots w_l$. To score $y_j$ given $m_i, c_i$, we first formulate the input as in Figure \ref{fig:prompt_recall}.
where $c_i^l, c_i^r$ are left and right context of $m_i$, and \texttt{`such as'} is the template we use to induce types. The input is then fed into BERT-large-uncased\footnote{We use the PLM from \url{https://huggingface.co}} for masked language model to obtain the probabilities of subwords, the score of $y_j$ is calculated by $ s^{MLM}_{j} = (\sum_{n=1}^l \log p_n)/l$, where $p_l$ denotes the probability of subword $w_l$ predicted by the PLM. We rank all types by enumerating all possible $l$ and recall $K_2$ additional candidates that haven't been recalled by the recall stage and exact match. We found that the expand stage improves recalls and contributes to the performance in Sec. \ref{sec:exp_expand}.
\subsection{Filter Stage}
In the filter stage, we use the recall and expand method introduced above to efficiently generate type candidates $\mathcal{C}_i$ for data in the train, development, and test set. For training, $\mathcal{C}_i$ is used to produce positive and hard negative type candidates. For inference, $\mathcal{C}_i$ is the candidate pool for the trained filter models. Let $|\mathcal{C}_i|=K$ and $K$ is typically less than 256. 
\subsubsection{CE} A trivial idea is to train a CE model introduced in Sec. \ref{sec:vanilla_ce} to filter $\mathcal{C}_i$ instead of filtering the whole type set $\mathcal{Y}$. The positive type $y^{+}$ and negative $y^{\_}$ type are both sampled from $\mathcal{C}_{i}$ and are used for calculating marginal ranking loss. To infer types, we also recall and expand $K$ candidates and score these candidates by $K$ forward passes to predict types. As $K << |\mathcal{Y}|$, CE with our Recall-Expand-Filter paradigm is much faster than vanilla CE. However, it's still inefficient compared to MLC-like models that concurrently predict scores of all types in a single forward pass. For faster inference and training, we propose multi-candidate cross-encoders ({\bf \textsc{\name}}) and introduce them in the next section.

\section{Multi-Candidate Cross-Encoder (\name)}
\label{sec:ccf}
 In this section, we introduce {\bf \textsc{\name}} for filtering candidates in one forward pass and propose several variants. 
\subsection{Overall Introduction of \name}
As shown in Figure \ref{fig:ccf}, compared to CE that concatenates one candidate at a time, {\bf \textsc{\name}} models concatenate all candidates in $\mathcal{C}_i$ with the mention and context. The input is then fed into the PLM to obtain the hidden states of each candidate as their representation. Finally, we use an MLP to concurrently score all candidates.
\begin{equation}
\begin{aligned}
\bm{h}_{1:K}  &= \texttt{PLM}(\texttt{[CLS]} \ c_i \ \texttt{[SEP]}\  m_i \ \texttt{[SEP]} \ t_{1:K} \ )  \\
\bm{s}_{1:K}  &= \texttt{Linear}(\bm{h}_{1:K}) 
\end{aligned}
\end{equation}
where $t_{1:K}$ is the short for $t_1, \cdots, t_K$, and $t_j \in \mathcal{C}_i$. Similarly, $\bm{h}_{1:K}$ and $s_{1:K}$ are hidden representations and scores of corresponding candidates respectively. 

\paragraph{Training and Inference} For training, we found that all positive types are ranked very high in the training candidates, which is not the case for the development and test data. To prevent the filter model from overfitting the order of training candidates and only learning to predict the first several candidates, we keep permuting type candidates during training. Same as the MLC model mentioned in Sec. \ref{sec:mlc}, we use the Binary Cross-Entropy loss as the training objective and tune a threshold of probabilities on the development set for inference.

In the next subsection, we discuss different model configurations of {\bf \textsc{\name}} regarding the input formats of candidates and attention mechanisms.

\subsection{Different Input Formats of Candidates}
\begin{figure}[h]
    \centering
    \scalebox{0.3}{
    \includegraphics{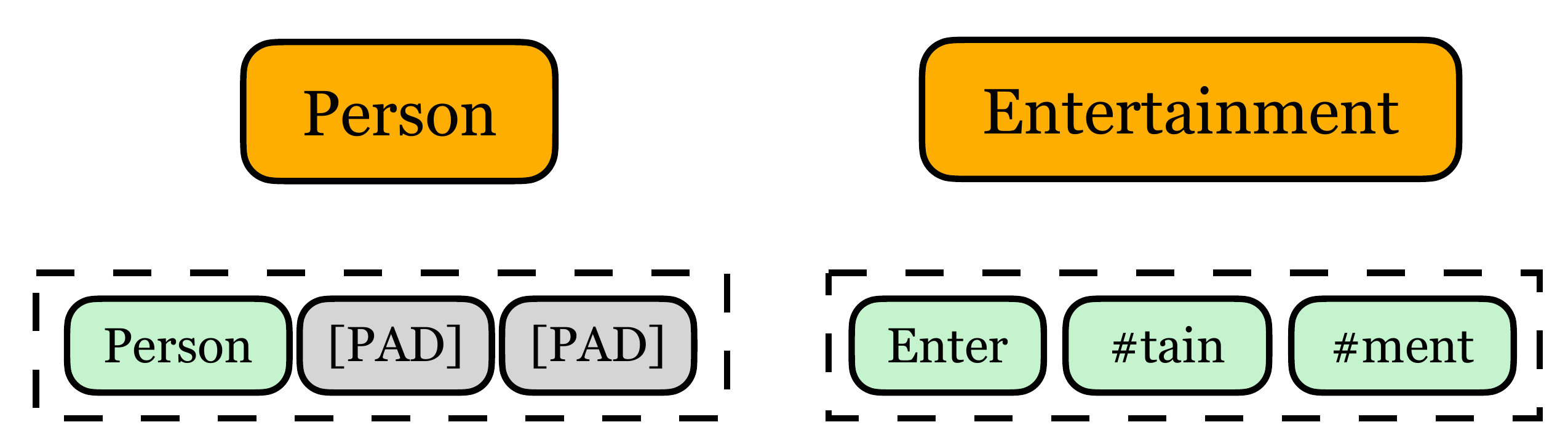}}
    \caption{Illustration of candidate block.}
    \label{fig:cand_block}
\end{figure}

\paragraph{Average of type sub-tokens} We treat each type $t_j \in \mathcal{Y}$ as a new token $u_j$ and add it to the vocabulary of PLM. The static embedding (layer 0 embedding of PLM) of $u_j$ is properly initialized by the average static embedding of $t_j$'s sub-tokens. As type candidates are capsuled into single tokens, the candidate representation $t_j$ is simply the last hidden state of $u_j$. The reasons for representing each type as a single token is (1) The max sequence length allowed by most PLMs is limited to 512, compressing types into single tokens is position saving. (2) Types in {\bf \textsc{UFET}} are tokenized into $2.1$ sub-tokens in average (by RoBERTa's tokenizer). Compressing types will not lose too many type semantics.
\paragraph{Fixed-size sub-token block} To preserve more type semantics, we place each candidate into a fixed-sized block as shown in Figure \ref{fig:cand_block}. We found the fixed block size makes PLM easier to enable the parallel implementation of different attention mechanisms that we will introduce next. We use the first hidden state in the block as the candidate representation.

\subsection{Attentions in \name}
\label{sec:attn}
\begin{figure}[h]
    \centering
    \scalebox{0.22}{
    \includegraphics{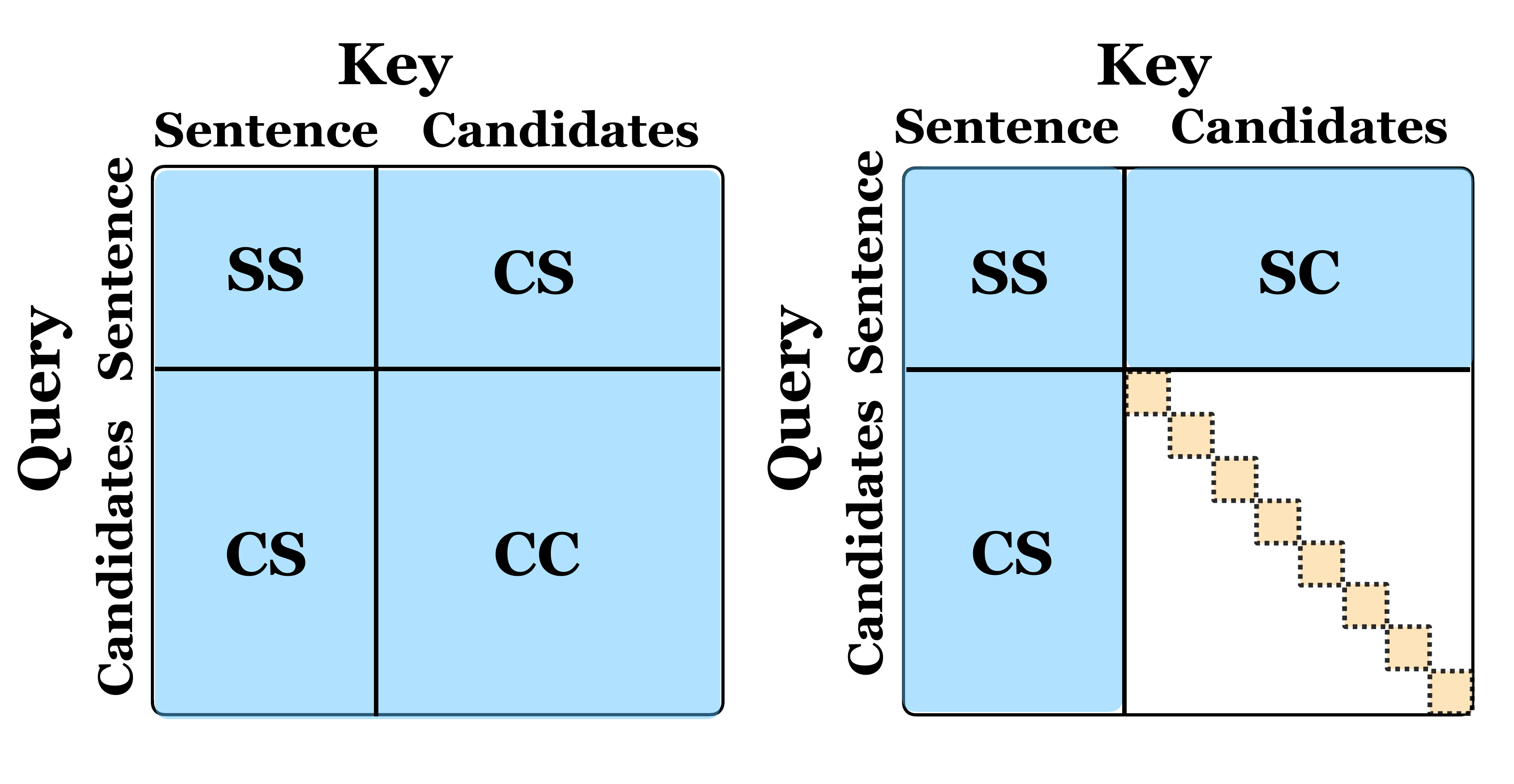}}
    \caption{Attentions in {\bf \textsc{\name}} (left), and {\bf \textsc{\name}}  without candidate-to-candidate (C2C) attention (right).}
    \label{fig:attn1}
\end{figure}

There are four kinds of attention in {\bf \textsc{\name}} as shown in Figure \ref{fig:attn1}, sentence to sentence (S2S), sentence to candidates (S2C), candidate to sentence (C2S), and candidate to candidate (C2C). As we score candidates based on mention and its context, attention from candidates to the sentence (C2S) is necessary. However, the necessity of C2C, S2S, and S2C is questionable. As our analytical experiment in Sec. \ref{sec:analyze} shows, it is important for words in the sentence to attend to all candidates (S2C), and is useful to have self-attention in the sentence (S2S), but the attentions (C2C) between different candidates are unnecessary. Based on these findings, we propose a new variant of {\bf \textsc{\name}} that the C2C attention is discarded in computation as shown in the right part of Figure \ref{fig:attn1}. Let $L_S$ and $L_C$ be the number of sub-tokens used by the sentence and candidates respectively. We can formulate the attention query of the sentence as $\bm{Q}_S=[\bm{q}^s_{1};\cdots;\bm{q}^s_{L_S}] \in \mathbb{R}^{L_S \times D}$,  where $\bm{q}^s_i$ is the query vector of the $i$-th sub-token in the sentence, and $D$ is the embedding dimension. Similarly, the query of candidates is formulated as $\bm{Q}_C=[\bm{q}^c_{1};\cdots;\bm{q}^c_{L_C}] \in \mathbb{R}^{L_C \times D}$. When we treat candidates as average of sub-tokens, $\bm{q}^c_i$ is a $D$-dimensional vector, and when we use fixed-sized blocks to place candidates, $\bm{q}^c_i \in \mathbb{R}^{B \times D}$ is the concatenation of the query vectors in the $i$-th candidate block and $B$ is the number of sub-tokens in a block. The keys and values are defined similarly as $\bm{K}_C, \bm{V}_C, \bm{O}_C \in \mathbb{R}^{L_C \times D},  \bm{K}_S, \bm{V}_S, \bm{O}_S \in \mathbb{R}^{L_S \times D}$. The attention outputs are computed as:
\begin{align}
& \bm{O}_S  = \text{Softmax} \big( \frac{\bm{Q}_S [\bm{K}_S; \bm{K}_C]^T}{\sqrt{D}} \big) \cdot [\bm{V}_S; \bm{V}_C] \\
&[\bm{A}_{CS}; \bm{A}_{CC}]  = \text{Softmax} \big( \frac{[\bm{Q}_C \bm{K}_S^T; \bm{M}_C^T]}{\sqrt{D}} \big)  \\
&\bm{M}_C  =  [{\bm{q}_1^c}^T \bm{k}_1^c; \cdots; {\bm{q}_{L_C}^c}^T \bm{k}_{L_C}^c] \\
&\bm{A}_{CC} = [\bm{a}^c_{1}; \cdots; \bm{a}^c_{L_C}] \\
&\bm{O}_C = \bm{A}_{CS} \bm{V}_S + \sum_{j=1}^{L_C} \bm{a}_j \bm{v}^c_j
\end{align}
where $\bm{A}_{CC}$ is the intra-candidate or intra-block attention, and $\bm{a}_j^c$ is a scaler when we treat candidates as average of sub-tokens and is a $B \times B$ matrix when we represent candidates as blocks. The last step (Eq. 8) can be parallelly implemented by Einstein summation.
In most cases, candidate length $L_C$ is significantly larger than sentence length $L_S$. As a result, by ignoring the C2C attention, the inference speed is further improved because the time complexity of the attention is significantly reduced from $O(D(L_S+L_C)^2)$ to $O(D(L_S^2+2L_SL_C+B^2L_C))$. More importantly, the space complexity in attention also gets reduced from $O((L_S+L_C)^2)$ to $O(L_S^2+2L_SL_C))$, which allows us to filter more candidates concurrently. The improvement in space and time complexity by discarding C2C attention is more obvious when the number of candidates becomes larger.

\section{Experiments}
We conduct experiments on two ultra-fine entity typing datasets, {\bf \textsc{UFET}} (English) and {\bf \textsc{CFET}} (Chinese). Their data statistics are shown in Table \ref{tab:stat}. We mainly focus on and report the macro-averaged recall at the recall and expand stage, and concern mainly on the macro-$F1$ of the final prediction at the filter stage. We also evaluate the {\bf \textsc{\name}} models on the fine-grained (130 types) and coarse-grained (9 types) settings of entity typing without the recall and expand stage.
\subsection{UFET and CFET}
\subsubsection{Recall Stage}
\label{sec:recall}
\begin{figure}[t]
     \centering
     \begin{subfigure}[h]{0.5\textwidth}
         \centering
         \includegraphics[width=\textwidth]{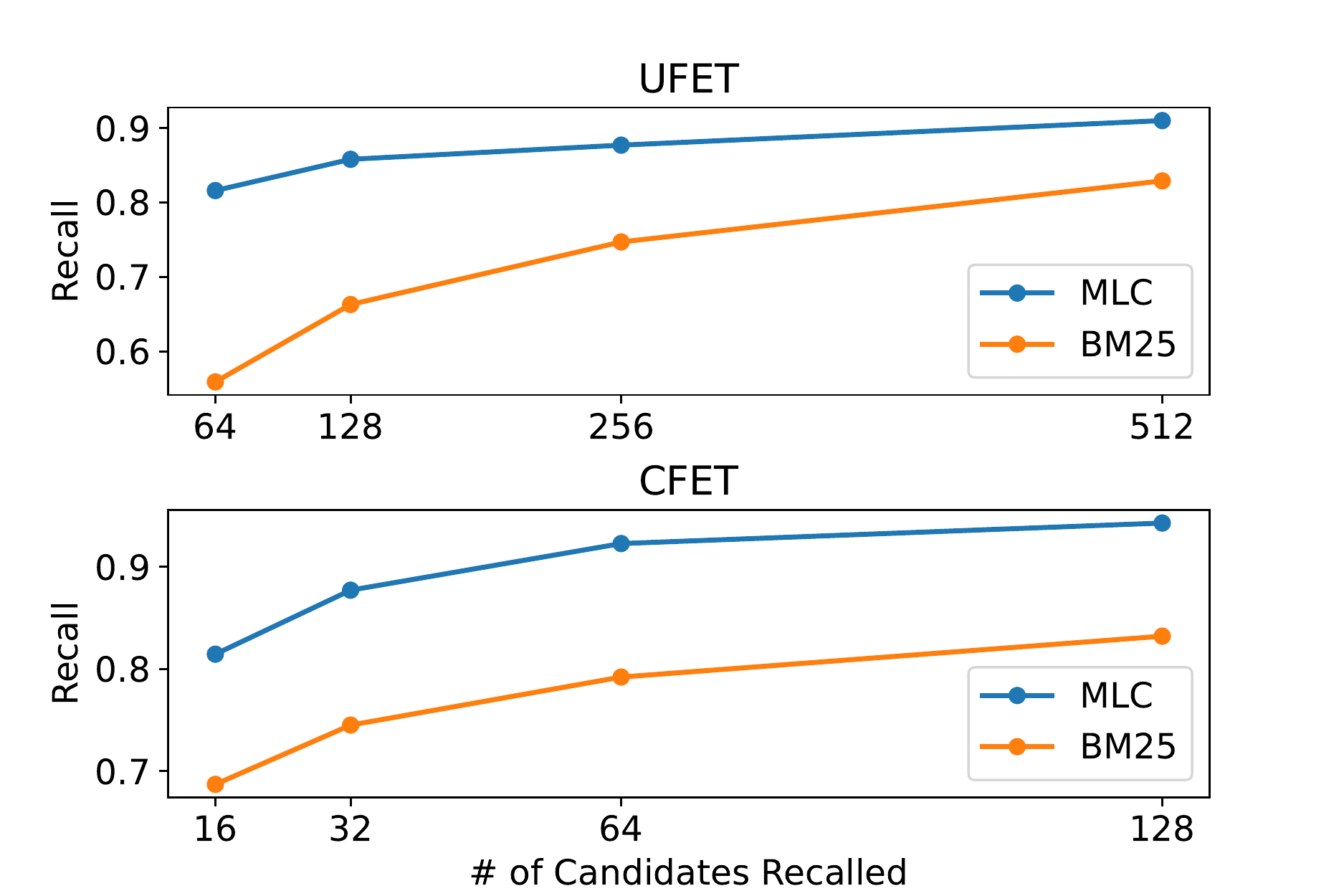}
         \label{fig:mb2}
     \end{subfigure}   
 \caption{Recall@$K$ of MLC and BM25.}
 \label{fig:recall}
\end{figure}

We compare the recall@$K$ on the test sets of {\bf \textsc{UFET}} and {\bf\textsc{CFET}} between the trained MLC model (introduced in \ref{sec:mlc}) and a traditional BM25 model \cite{bm25} in Figure \ref{fig:recall}. The MLC model uses the RoBERTa-large as backbone and is tuned based on the recall@$128$ on the development set. We use AdamW optimizer with a learning rate of $2\times10^{-5}$. Results show that MLC is a strong recall model, it consistently has better recall compared to BM25 on both {\bf\textsc{UFET}} and {\bf\textsc{CFET}} dataset, and the recall@$128$ reaches over $85\%$ on {\bf \textsc{UFET}}, and over $94\%$ on {\bf \textsc{CFET}}.

\subsection{Expand Stage}
\label{sec:expand}
\begin{figure}[t]
     \centering
     \begin{subfigure}[h]{0.45\textwidth}
         \centering
         \includegraphics[width=\textwidth]{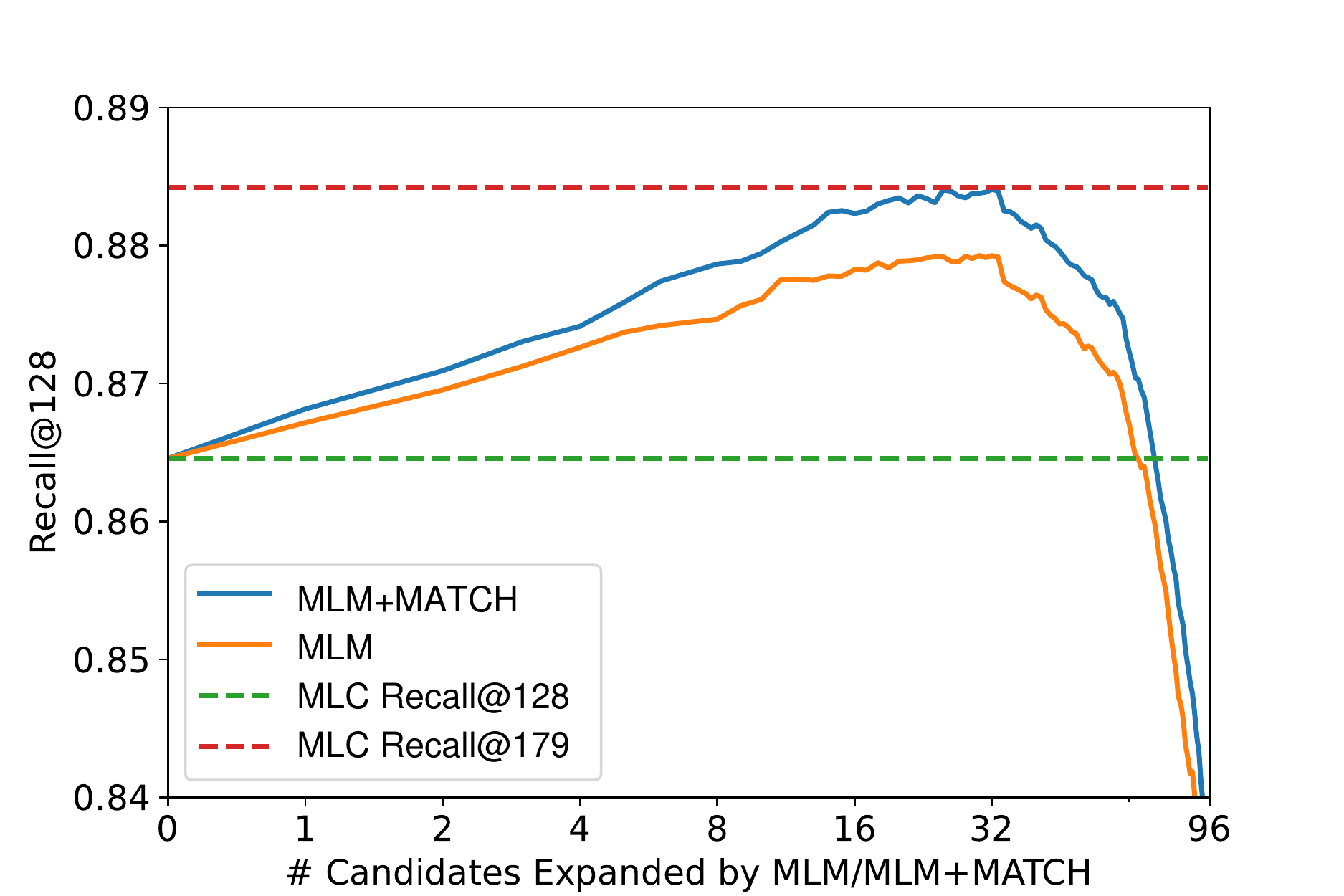}
         \caption{Recall@$128$ on {\bf \textsc{UFET}} by including different number of expanded candidates. }
         \label{fig:c1}
     \end{subfigure}
     \vfill
     \begin{subfigure}[h]{0.45\textwidth}
         \centering
         \includegraphics[width=\textwidth]{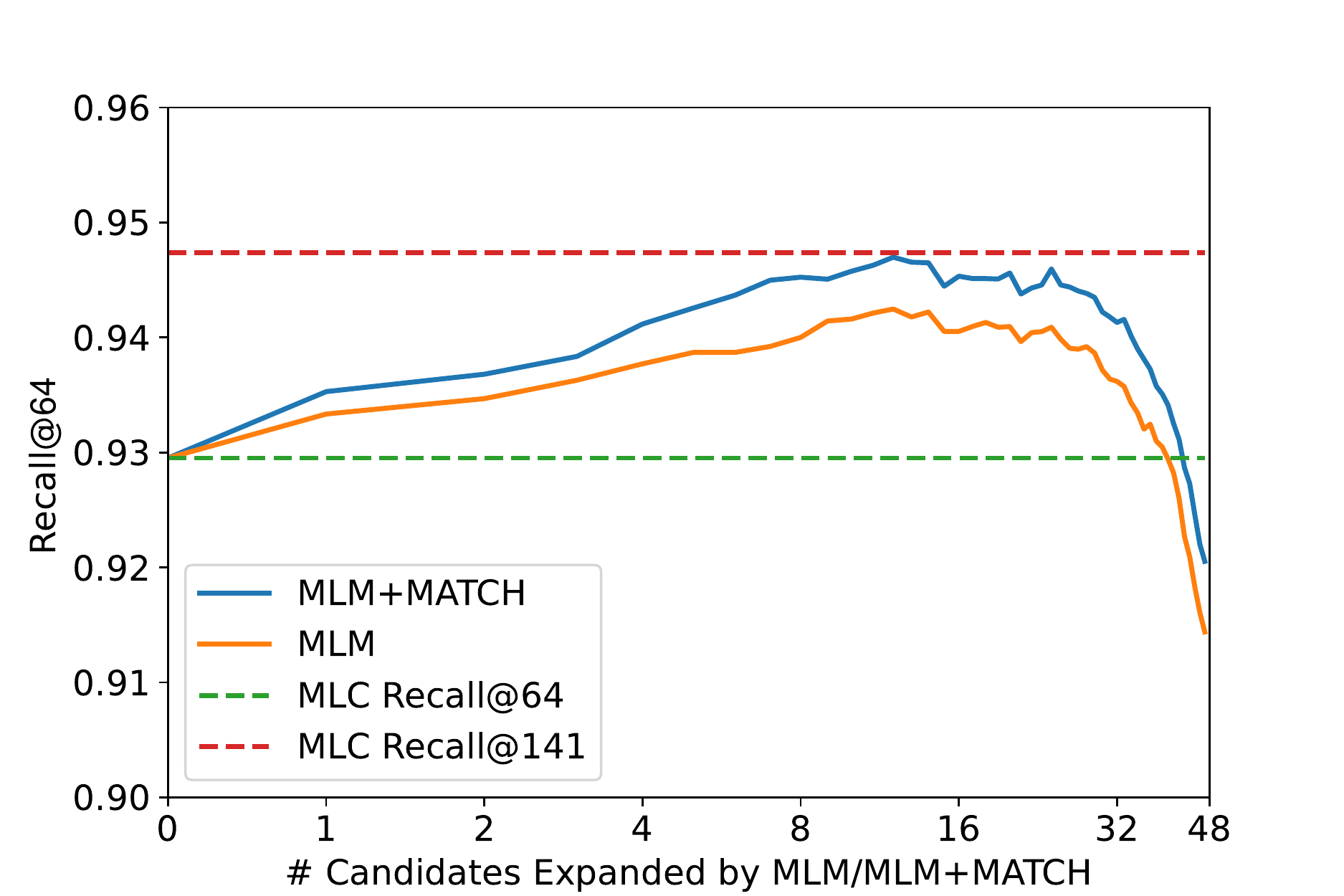}
         \caption{Recall@$64$ on {\bf \textsc{CFET}} by including different number of expanded candidates.}
         \label{fig:c2}
     \end{subfigure}
\caption{Demonstration of the effect of expand stage. $x$-axis represents the number of candidates expanded by MLM/MLM+MATCH among these $128$ candidates. }
\label{fig:expand_improvement}
\end{figure}

In Table \ref{tab:expand}, we evaluate the F1 scores of all candidates expanded by exact match, and top-$10$ candidates expanded by the MLM using Bert-large. We also demonstrate the improvement of recall by using candidate expansion in Figure \ref{fig:expand_improvement}. On {\bf \textsc{UFET}} dataset, expanding around $32$ additional candidates based on $112$ MLC candidates results in $2\%$ higher recall compared to recalling all $128$ candidates by MLC. The recall of $128$ candidates after the expansion is comparable to the recall of $180$ candidates recalled from MLC. Similarly, expanding $10$ candidates is comparable to additionally recalling $80$ candidates using MLC.
In our experiments, we replace the last $48$ candidates recalled by MLC with the candidates recalled by MLM and Exact match for {\bf \textsc{UFET}} and $10$ for {\bf \textsc{CFET}}. We found the expand stage has a positive effect on the final performance of {\bf \textsc{\name}}s, and helps them reach SOTA performance (analyze in Sec. \ref{sec:analyze}).

\begin{table}[t]
\centering
\scalebox{0.75}{
\begin{tabular}{cccccc} 
\toprule
{\bf \textsc{Dataset}} & {\bf \textsc{Expand}} &   {\bf \textsc{P}}  & {\bf \textsc{R}}  &  {\bf \textsc{F1}} & \small{Avg \# Expanded}  \\ \midrule
\multirow{2}{*}{\bf \textsc{UFET}} & {\bf \textsc{Match}}      & 11.2   & 11.3     & 9.8    & 5.23     \\
      & {\bf \textsc{MLM}}  &  8.5     &   17.1   &  10.7  &    10    \\ \midrule
\multirow{2}{*}{\bf \textsc{CFET}} & {\bf \textsc{Match}}   &  11.4  &  14.5  & 11.2   & 4.57    \\
 & {\bf \textsc{MLM}}  & 21.3   &  19.5  & 17.7    & 10    \\ \midrule
\end{tabular}}
\caption{Evaluation of the recalled candidates.}
\label{tab:expand}
\end{table}
\label{sec:exp_expand}
\subsection{Filter Stage and Final Results.}
\begin{table}[h!]
\centering
\scalebox{0.73}{
\renewcommand{\arraystretch}{1}
\begin{tabular}{cllll} \toprule
\multicolumn{2}{l}{\bf \textit{Base Models on UFET} }     & \bf \textsc{P}    & \bf \textsc{R}   & \bf \textsc{F1}  \\ \midrule
\multicolumn{5}{l}{\emph{MLC-like models}}        \\
\color{blue} \bf \texttt{B}& {\bf \textsc{Box4Types}}\cite{box4types}  & 52.8 & 38.8 & 44.8  \\
\color{blue}\bf \texttt{B}& {\bf \textsc{LDET}}$^\dagger$  \cite{onoe-durrett-2019-learning}          & 51.5 & 33.0 & 40.1 \\ 
\color{blue}\bf \texttt{B}& {\bf \textsc{MLMET}}$^\dagger$   {\cite{mlmet}}   & 53.6 & 45.3 & 49.1  \\
\color{blue}\bf \texttt{B}& {\bf \textsc{PL}}  \cite{ding2021prompt}   & 57.8 & 40.7 & 47.7 \\
\color{blue}\bf \texttt{B}& {\bf \textsc{DFET}}    \cite{dfet}      & 55.6 & 44.7 & 49.5 \\
\color{blue}\bf \texttt{B}& {\bf \textsc{MLC}} (reimplemented by us) & 46.5 & 34.9 & 39.9 \\ 
\color{red}\bf \texttt{R}& {\bf \textsc{MLC}} (reimplemented by us) & 42.2 & 44.9 & 43.5 \\ \hline 
\multicolumn{5}{l}{\emph{Seq2seq based models}}      \\
\color{blue}\bf \texttt{B} & {\bf \textsc{LRN} }  {\cite{liu-etal-2021-fine}}              & 54.5 & 38.9 & 45.4  \\\hline
\multicolumn{5}{l}{\emph{Filter models under our recall-expand-filter paradigm}}      \\
\color{blue}\bf \texttt{B} & {\bf \textsc{Vanilla CE}$_{128}$}   & 47.2 & 48.5 & 47.8 \\ 
\color{blue}\bf \texttt{B} & {\bf \textsc{\name-S$_{128}$}} (Ours)  & 53.2 & 48.3 & {\bf 50.6} \\ 
\color{blue}\bf \texttt{B} & {\bf \textsc{\name-S$_{128}$ w/o C2C}}   (Ours)   & 52.3 & 48.3 & 50.2 \\ 
\color{blue}\bf \texttt{B} & {\bf \textsc{\name-B$_{128}$}} (Ours)    & 49.9 & 50.0 & 49.9 \\ 
\color{blue}\bf \texttt{B} & {\bf \textsc{\name-B$_{128}$ w/o C2C}} (Ours)     & 49.9 & 48.2 & 49.0 \\ \hline
\color{red}\bf \texttt{R} & {\bf \textsc{Vanilla CE}$_{128}$}   & 49.6 & 49.0 & 49.3 \\ 
\color{red}\bf \texttt{R} & {\bf \textsc{\name-S$_{128}$}} (Ours)  & 53.3 & 47.3 & 50.1 \\ 
\color{red}\bf \texttt{R} & {\bf \textsc{\name-S$_{128}$ w/o C2C}}   (Ours)  & 53.2 & 46.6 & 49.7 \\ 
\color{red}\bf \texttt{R} & {\bf \textsc{\name-B$_{128}$}} (Ours)  & 52.5 & 47.9 & 50.1 \\ 
\color{red}\bf \texttt{R} & {\bf \textsc{\name-B$_{128}$ w/o C2C}} (Ours)     & 52.7 & 46.4 & 49.3 \\ \hline
\midrule
\multicolumn{2}{l}{\bf \textit{Large Models on UFET} }     & \bf \textsc{P}    & \bf \textsc{R}   & \bf \textsc{F1}  \\ \midrule
\multicolumn{5}{l}{\emph{MLC-like models}}        \\
\color{red}\bf \texttt{R} & {\bf \textsc{MLC}}  \cite{npcrf}               & 47.8 & 40.4 & 43.8  \\
\color{red}\bf \texttt{R} & {\bf \textsc{MLC-NPCRF}} \cite{npcrf}             & 48.7 & 45.5 & 47.0  \\
\color{red}\bf \texttt{R} & {\bf \textsc{MLC-GCN}} \cite{xiong-etal-2019-imposing}     & 51.2 & 41.0 & 45.5 \\
\color{blue}\bf \texttt{B} & {\bf \textsc{PL}}  \cite{ding2021prompt}       & 59.3 & 42.6 & 49.6  \\
\color{blue}\bf \texttt{B} & {\bf \textsc{PL-NPCRF}}  \cite{npcrf}  & 55.3 & 46.7 & {50.6}\\ \hline
\multicolumn{4}{l}{\emph{Cross-encoder based models and {\bf \textsc{\name}}s}}      \\
\color{red}\bf \texttt{R} & {\bf \textsc{LITE+L}}  \cite{lite}             & 48.7 & 45.8 & 47.2  \\
\color{teal}\bf \texttt{RM} & {\bf \textsc{LITE+NLI+L}} \cite{lite} & 52.4 & 48.9 & {50.6} \\ \hline
\multicolumn{4}{l}{\emph{Filter models under our recall-expand-filter paradigm}}   \\ 
\color{blue}\bf \texttt{B} & {\bf \textsc{Vanilla CE$_{128}$}}   & 50.3 & 49.6 & 49.9 \\ 
\color{blue}\bf \texttt{B} & {\bf \textsc{\name-S$_{128}$}}  (Ours)   & 52.5 & 49.1 & 50.8 \\ 
\color{blue}\bf \texttt{B} & {\bf \textsc{\name-S$_{128}$ w/o C2C}}   (Ours)   & 54.1 & 47.1 & 50.4 \\ 
\color{blue}\bf \texttt{B} & {\bf \textsc{\name-B$_{128}$}} (Ours)    & 54.0 & 48.6 & 51.2 \\ 
\color{blue}\bf \texttt{B} & {\bf \textsc{\name-B$_{128}$ w/o C2C}} (Ours)     & 52.8 & 48.3 & 50.4 \\ \hline
\color{red}\bf \texttt{R} & {\bf \textsc{Vanilla CE$_{128}$}}   & 54.5 & 49.3 & 51.8 \\ 
\color{red}\bf \texttt{R} & {\bf \textsc{\name-S$_{128}$}}  (Ours)   & 50.8 & 49.8  &  50.3 \\ 
\color{red}\bf \texttt{R} & {\bf \textsc{\name-S$_{128}$ w/o C2C}}   (Ours)   & 51.5 & 48.8 & 50.1 \\ 
\color{red}\bf \texttt{R} & {\bf \textsc{\name-B$_{128}$}} (Ours)    & 51.9 & 50.8 & 51.4 \\ 
\color{red}\bf \texttt{R} & {\bf \textsc{\name-B$_{128}$ w/o C2C}} (Ours)     & 51.6 & 51.6 & 51.6 \\ \hline
\color{teal}\bf \texttt{RM} & {\bf \textsc{\name-B$_{128}$ w/o C2C}} (Ours) & 56.3 & 48.5 & {\bf 52.1} \\ \hline
\midrule
\end{tabular}}
\caption{Macro-averaged UFET result. {\bf \textsc{LITE+L}} is LITE without NLI pretraining, {\bf \textsc{LITE+L+NLI}} is the full LITE model. Methods marked by $\dagger$ utilize either distantly supervised or augmented data for training. {\bf \textsc{\name-S$_{128}$}} denotes we use $128$ candidates recalled and expanded from the first two stages.}
\label{tab:ufet}
\end{table}
\begin{table}[t]
\centering
\scalebox{0.75}{
\renewcommand{\arraystretch}{1}
\begin{tabular}{cllll} \toprule
\multicolumn{2}{l}{\bf \textit{Models on CFET} }     & \bf \textsc{P}    & \bf \textsc{R}   & \bf \textsc{F1}  \\ \midrule
\multicolumn{5}{l}{\emph{MLC-like models}}        \\
\color{purple}\bf \texttt{N}& {\bf \textsc{MLC}} & 55.8 & 58.6 & 57.1 \\  
\color{purple}\bf \texttt{N}& {\bf \textsc{MLC-NPCRF}} \cite{npcrf}     & 57.0 & 60.5 & 58.7 \\ 
\color{purple}\bf \texttt{N}& {\bf \textsc{MLC-GCN}} \cite{xiong-etal-2019-imposing}   & 51.6 & 63.2 & 56.8 \\ 
\color{brown}\bf \texttt{C}& {\bf \textsc{MLC}} & 54.0 & 59.5 & 56.6 \\  
\color{brown}\bf \texttt{C}& {\bf \textsc{MLC-NPCRF}} \cite{npcrf}   & 54.0 & 61.6 & 57.3 \\  
\color{brown}\bf \texttt{C}& {\bf \textsc{MLC-GCN}} \cite{xiong-etal-2019-imposing} & 56.4 & 58.6 & 57.5 \\ \midrule 
\multicolumn{5}{l}{\emph{Filter models under our recall-expand-filter paradigm}}      \\
\color{purple}\bf \texttt{N} & {\bf \textsc{Vanilla CE}}   & 57.6 & 64.3 & 60.7 \\ 
\color{brown}\bf \texttt{C} & {\bf \textsc{Vanilla CE}}   & 54.0 & 63.3 & 58.3 \\  \hline
\color{purple}\bf \texttt{N} & {\bf \textsc{\name-S$_{64}$}} (Ours)  & 58.4 & 62.1 & 60.2 \\ 
\color{purple}\bf \texttt{N} & {\bf \textsc{\name-S$_{64}$ w/o C2C}}   (Ours)   & 59.1 & 61.5 & 60.3 \\ 
\color{purple}\bf \texttt{N} & {\bf \textsc{\name-B$_{64}$}} (Ours)    & 56.7 & 66.1 & 61.1 \\ 
\color{purple}\bf \texttt{N} & {\bf \textsc{\name-B$_{64}$ w/o C2C}} (Ours)     & 58.8 & 64.1 & 61.4 \\ \hline
\color{brown}\bf \texttt{C} & {\bf \textsc{\name-S$_{64}$}} (Ours)  & 55.5 & 62.6 & 58.8 \\ 
\color{brown}\bf \texttt{C} & {\bf \textsc{\name-S$_{64}$ w/o C2C}}   (Ours)   & 54.0 & 63.4 & 58.3 \\ 
\color{brown}\bf \texttt{C} & {\bf \textsc{\name-B$_{64}$}} (Ours)    & 55.0 & 63.5 & 59.0 \\ 
\color{brown}\bf \texttt{C} & {\bf \textsc{\name-B$_{64}$ w/o C2C}} (Ours)     & 57.3 & 61.3 & 59.3 \\ \hline
\midrule
\end{tabular}}
\caption{Macro-averaged CFET result.}
\label{tab:cfet}
\end{table}

In this section, we report the performance of {\bf \textsc{MCCE}} variants as the filter models and compare them with various strong baselines that we will introduce later. We also compare the inference speed of different models in this section. For filter models, we treat the number of candidates $K$ recalled and expanded by the first two stages as hyper-parameters, and tune it on the development set. We found the choice of PLM backbones has a non-negligible effect on the performance, and the PLM backbone of previous methods varies. Therefore for fairer comparisons to baselines, we conduct experiments of {\bf \textsc{\name}} using different backbone PLMs for our {\bf \textsc{\name}} models and report the results. For all {\bf \textsc{\name}} models, we use AdamW optimizer with a learning rate tuned between $5\times 10^{-6}$ and $2\times 10^{-5}$. The batch size we use is $4$ and we train the models for at most $50$ epochs with early stopping. {\bf \textsc{UFET}} also provides a large dataset obtained from distant supervision such as entity linking, we do not use it and only train and evaluate our models on human-labeled data.
\paragraph{Baselines}
The {\bf \textsc{MLC}} model we used for the recall stage and the cross-encoder ({\bf \textsc{CE}}) we introduced in Sec. \ref{sec:vanilla_ce} are natural baselines. We also compare our methods with recent PLM-based methods. {\bf \textsc{LDET} }\cite{onoe-durrett-2019-learning} is an MLC with Bert-base-uncased and ELMo \cite{elmo} trained on 727k examples automatically denoised from the distantly labeled UFET. {\bf \textsc{GCN} }\cite{xiong-etal-2019-imposing} uses GCN to model type correlations and obtain type embeddings. Types are scored by dot-product of mention and type embeddings. The original paper uses BiLSTM as the mention encoder and we use the results re-implemented by \citet{npcrf} using RoBERTa-large. {\bf \textsc{Box4Type} }\cite{box4types} uses Bert-large as the backbone and uses box embedding to encode mentions and types for training and inference. {\bf \textsc{LRN} }\cite{liu-etal-2021-fine} use Bert-base as the encoder and an LSTM decoder to generate types in a seq2seq manner. {\bf \textsc{MLMET} }\cite{mlmet} is a {\bf \textsc{MLC}} with Bert-base, but first pretrained by the distantly-labeled data augmented by masked word prediction, then finetuned and self-trained on the 2k human-annotated data. {\bf \textsc{PL}} \cite{ding2021prompt} uses prompt learning for entity typing. {\bf \textsc{DFET} }\cite{dfet} uses {\bf \textsc{PL}} as backbone and is a multi-round automatic denoising method for 2k labeled data. {\bf \textsc{LITE} }\cite{lite} is the previous SOTA system that formulates entity typing as textual inference. {\bf \textsc{LITE}} uses RoBERTa-large-MNLI as the backbone, and is a cross-encoder (introduced in Sec. \ref{sec:vanilla_ce}) with designed templates and a hierarchical loss. \citet{npcrf} proposes {\bf \textsc{NPCRF}} to enhance backbones such as {\bf \textsc{PL}} and {\bf \textsc{MLC}} by modeling type correlations, and reach performance comparable to {\bf \textsc{LITE}}.

\paragraph{Naming Conventions}
Let {\bf \textsc{\name-S}} be the {\bf \textsc{\name}} model that treats candidates as sub-tokens, and {\bf \textsc{\name-B}} be the model representing candidates as fixed-size blocks. The {\bf \textsc{\name}} model without {\bf \textsc{C2C}} attention (mentioned in Sec. \ref{sec:attn}) is denoted as {\bf \textsc{\name-B} w/o C2C}. For PLM backbones used in {\bf \textsc{UFET}}, we use {\color{blue} \bf \texttt{B}}, {\color{red} \bf \texttt{R}}, {\color{teal} \bf \texttt{RM}} to denote BERT-base-cased \cite{bert}, RoBERTa \cite{liu2019roberta}, and RoBERTa-MNLI \cite{liu2019roberta} respectively. For {\bf \textsc{CFET}}, we adopt two widely-used Chinese PLM, BERT-base-Chinese and NeZha-base-Chinese, and denote them as {\color{brown} \bf \texttt{C}} and {\color{purple} \bf \texttt{N}} respectively. 

\paragraph{UFET Results} We show the results of {\bf \textsc{UFET}} dataset in Table \ref{tab:ufet}. The results show that: (1) The recall-expand-filter paradigm is effective. Filter models outperform all baselines without the paradigm by a large margin. The vanilla CE under our paradigm reaches $51.8$ F1 compared to more complexed CE {\bf \textsc{LITE}} with $50.6$ F1 (2) {\bf \textsc{\name}} models reach SOTA performances. {\bf \textsc{\name-S$_{128}$}} with BERT-base performs best and reaches {\bf 50.6} F1 score, which is comparable to previous SOTA performance of large models such as {\bf \textsc{LITE+NLI+L}} and {\bf \textsc{PL+NPCRF}}. Among large models, {\bf \textsc{\name-B$_{128}$ w/o C2C}} also reaches SOTA performance with {\bf 52.1} F1 score. (3) {\bf \textsc{C2C}} attention is not necessary on large models, but is useful in base models. (4) Large models can utilize type semantics better. We found {\bf \textsc{\name-B}} outperforms {\bf \textsc{\name-S}} on large models, but underperforms {\bf \textsc{\name-S}} on base models. (5) Backbone PLM matters. We found the performance of {\bf \textsc{Vannila CE}} under our paradigm is largely affected by the PLM it used. It reaches $47.8$ F1 with BERT-base and $51.8$ F1 with RoBERTa-large. For {\bf \textsc{\name}} models, we found {\bf \textsc{\name}} performs better than {\bf \textsc{\name-B}} with BERT, and worse than {\bf \textsc{\name-B}} with RoBERTa. 

\begin{table*}[t]
\centering
\scalebox{0.9}{
\begin{tabular}{lllcc} \toprule
\bf \textsc{Model}  & \bf \textsc{\# FP} & \bf \textsc{Attn} & \bf \textsc{sents/sec} & \bf \textsc{F1} \\ \midrule
{\bf \textsc{MLC}} & \small{$1$}  & \small{$L_S^2D$} & 58.8 & 43.8\\
{\bf \textsc{LITE+NLI+L (CE)}}  & \small{$N$}  & \small{$L_S^2D$} & 0.02 & 50.6\\ \midrule \hline
\multicolumn{5}{l}{\emph{filter stage inference speed.}}  \\
{\bf \textsc{Vanilla CE$_{128}$}}  & \small{$128$}  & \small{$L_S^2D$} & 1.64 & 51.8 \\ 
{\bf \textsc{\name-S$_{128}$}}  & \small{$1$}  & \small{$(L_S+128)^2D$} & 60.8 & 50.1 \\ 
{\bf \textsc{\name-B$_{128}$}}  & \small{$1$}  & \small{$(L_S+128B)^2D$} & 22.3 & 51.4\\ 
{\bf \textsc{\name-B$_{128}$ w/o C2C}}  & \small{$1$}  & \small{$(L_S^2+256L_S B + 128 B^2)D$} & 25.2 & {\bf 52.1}\\ \bottomrule
\end{tabular}}
\caption{Inference speed comparison of models. {\bf \textsc{\# FP}} means the number of PLM forward passes required by a single inference. {\bf \textsc{ATTN}} column lists the theoretical attention complexity.  We also report the practical inference speed {\bf \textsc{sents/sec}} and the {\bf \textsc{F1}} scores on {\bf \textsc{UFET}} with RoBERTa-large architecture.}
\label{tab:speed}
\end{table*}

\begin{table}[t]
\centering
\scalebox{0.85}{
\renewcommand{\arraystretch}{1}
\begin{tabular}{cllll} \toprule
\multicolumn{2}{l}{\bf \textit{Models} }     & \bf \textsc{P}    & \bf \textsc{R}   & \bf \textsc{F1}  \\ \midrule
\multicolumn{5}{l}{\emph{coarse (9 types) Open Entity}}        \\ \hline
\color{red}\bf \texttt{R} & {\bf \textsc{MLC}}   & 76.8 & 78.5 & 77.6 \\ 
\color{red}\bf \texttt{R} & {\bf \textsc{Vanilla CE$_{9}$}}   & 82.3 & 81.0 & 81.6 \\ 
\color{red}\bf \texttt{R} & {\bf \textsc{\name-S$_{9}$}}   & 77.0 & 87.7 & 82.0 \\ 
\color{red}\bf \texttt{R} & {\bf \textsc{\name-B$_{9}$ w/o C2C}}   & 77.2 & 85.4 & 81.1 \\ \hline
\multicolumn{5}{l}{\emph{fine (130 types)}}        \\ \hline
\color{red}\bf \texttt{R} & {\bf \textsc{MLC}}   & 70.4 & 63.7 & 66.9  \\ 
\color{red}\bf \texttt{R} & {\bf \textsc{Vanilla CE}$_{130}$}   & 67.9 & 66.4 & 67.1 \\ 
\color{red}\bf \texttt{R} & {\bf \textsc{\name-S$_{130}$}}   & 65.8 & 71.8 & 68.7 \\ 
\color{red}\bf \texttt{R} & {\bf \textsc{\name-B$_{130}$ w/o C2C}}   & 64.1 & 70.5 & 67.1 \\ \hline
\midrule
\end{tabular}}
\caption{Micro-averaged results on UFET fine and coarse.}
\label{tab:ufet-coarse-fine}
\end{table}

\paragraph{CFET Results} We conduct experiments on {\bf \textsc{CFET}} and compare {\bf \textsc{\name}} models with several strong baselines:  {\bf \textsc{NPCRF}} and {\bf \textsc{GCN}} with MLC-like architecture, and {\bf \textsc{Vanilla CE}} under out paradigm which is proved to be better than {\bf \textsc{LITE}} on {\bf \textsc{UFET}}. The results are shown in Table \ref{tab:cfet}. Similar to results in {\bf \textsc{UFET}}, filter models under our paradigm significantly outperform MLC-like baselines, $+2.0$ F1 for Nezha-base and $+1.8$ F1 for BERT-base-Chinese. In {\bf \textsc{CFET}}, {\bf \textsc{\name}-B} is significantly better than {\bf \textsc{\name}-S}, on both Nezha-base and BERT-base-Chinese, indicating the importance of type semantics in Chinese language. We also find that {\bf \textsc{\name} w/o C2C} is generally better than  {\bf \textsc{\name} w/ C2C}, it is possibly because the C2C attention distracts the candidates from attending to mention and contexts.
\paragraph{Speed Comparison} Table \ref{tab:speed} shows the theoretical inference complexity (number of PLM forward passes, and attention complexity), and practical inference speed (number of sentences inferred per second) of different models. We conduct the speed test using NVIDIA TITAN RTX for all models, and the inference batch size is 4.
At the filter stage, the inference speed of {\bf \textsc{\name-S}} is on par with {\bf \textsc{MLC}} (even slightly faster because we don't need to score all types), and is about 40 times faster than {\bf \textsc{Vannila CE}} and thousands of times faster than {\bf \textsc{LITE}}. {\bf \textsc{\name-B w/o C2C}} is not significantly faster than {\bf \textsc{\name-B}} as expected. It's possibly because the computation related to the block attention is not fully optimized by existing deep learning frameworks. The speed advantage of {\bf \textsc{\name-B w/o C2C}} over {\bf \textsc{\name-B}} will be greater with more candidates.

\subsection{Fine-grained and Coarse-grained Entity Typing}
We also conduct experiments on Fine-grained (130-class) and Coarse-grained (9-class, also known as ``Open Entity'') entity typing, and the results are shown in Table \ref{tab:ufet-coarse-fine}. As the type candidate set is much smaller in these settings, we skip the recall and expand stages and directly run the filter models and compare them to baselines. Results show that both {\bf \textsc{\name}-S} and {\bf \textsc{\name}-B} are still better than {\bf \textsc{MLC}} and {\bf \textsc{Vanilla CE}}, and {\bf \textsc{\name}-S} is better than {\bf \textsc{\name}-B} on coarser-grained cases possibly because the coarser-grained types are simpler in surface-forms and {\bf \textsc{\name}-S} will not lose many type semantics.

\section{Analysis}
\label{sec:analyze}
\subsection{Importance of Expand Stage}
We perform the ablation study on the importance of the expand stage and show the results in Table \ref{fig:ablation_expand}. We compare the performances of {\bf \textsc{\name-S}} using the expanded or the not expanded candidate sets on {\bf \textsc{UFET}} and {\bf \textsc{CFET}}. We replace the last $48$ candidates recalled by MLC with candidates expanded by MLM and exact matching for {\bf \textsc{UFET}}, and $10$ candidates for {\bf \textsc{CFET}}. Results show that expand stage has a positive effect on performance, it improves the final recall by $+1.0$ and $+2.2$ on  {\bf \textsc{UFET}} and {\bf \textsc{CFET}} without harming the precision.

\begin{table}[t]
\centering
\scalebox{0.75}{
\renewcommand{\arraystretch}{1}
\begin{tabular}{cllll} \toprule
\multicolumn{2}{l}{\bf \textit{Ablation of Expand Stage} }     & \bf \textsc{P}    & \bf \textsc{R}   & \bf \textsc{F1}  \\ \midrule
\multicolumn{5}{l}{\bf \textsc{UFET\ \  MCCE with C2C BERT-large}} \\
\color{blue}\bf \texttt{B} & {\bf \textsc{\name-S$_{128}$ }} (Ours)     & 52.5 & 49.1 & 50.8 \\ 
\color{blue}\bf \texttt{B} & {\bf \textsc{\name-S$_{128}$ w/o Expand }} (Ours)     & 52.7 & 48.1 & 50.3\\ \hline
\multicolumn{5}{l}{\bf \textsc{CFET\ \  MCCE with C2C BERT-base-Chinese}} \\
\color{brown}\bf \texttt{C} & {\bf \textsc{\name-S$_{64}$}} (Ours)  & 55.5 & 62.6 & 58.8 \\ 
\color{brown}\bf \texttt{C} & {\bf \textsc{\name-S$_{64}$ w/o Expand}}   (Ours)   & 55.4 & 60.4 & 57.8 \\ \hline
\midrule
\end{tabular}}
\caption{Ablation study of expand stage.}
\label{fig:ablation_expand}
\end{table}

\subsection{Attentions}
We conduct an ablation study on S2S, C2S, S2C, and C2C attention introduced in Sec. \ref{sec:attn} and show the results in Table \ref{tab:attn}. From the results, we are surprised to find that removing C2C and S2S doesn't have a big negative impact on performance. The {\bf \textsc{\name-S}} using BERT-base reaches $48.8$ F1 even without both C2C and S2S attention. One possible reason is that the interaction between sub-tokens in the sentence can be achieved indirectly by first attending to the candidates and then being attended back by the candidate in the next layer. We also find that the C2S is necessary for the task ($18.7$ F1 without C2S) because we rely on the mention and its context to encode and classify candidates. Furthermore, we found that it is important for sentences to attend to all candidates (S2C), indicating that the interaction between the sentence and different types is crucial for the task.

\begin{table}[t]
\centering
\scalebox{0.75}{
\renewcommand{\arraystretch}{1}
\begin{tabular}{cllll} \toprule
\multicolumn{2}{l}{\bf \textit{Analysis about attention on UFET}}     & \bf \textsc{P}    & \bf \textsc{R}   & \bf \textsc{F1}  \\ \midrule
\multicolumn{5}{l}{\bf \textsc{\name-S using BERT-base}} \\
\color{blue}\bf \texttt{B} & {\bf \textsc{\name-S$_{128}$} FULL}     & 53.2 &  48.3 & 50.6 \\ 
\color{blue}\bf \texttt{B} & {\bf \textsc{\name-S$_{128}$ w/o C2C }}     & 52.3 & 48.3 & 50.2 \\
\color{blue}\bf \texttt{B} & {\bf \textsc{\name-S$_{128}$ w/o S2S }}     & 50.6 & 48.4 & 49.4 \\
\color{blue}\bf \texttt{B} & {\bf \textsc{\name-S$_{128}$ w/o S2C }}     & 48.7 & 47.1 & 47.9 \\ 
\color{blue}\bf \texttt{B} & {\bf \textsc{\name-S$_{128}$ w/o C2S }}     & 19.7 & 17.4 & 18.7\\
\color{blue}\bf \texttt{B} & {\bf \textsc{\name-S$_{128}$ w/o S2S,C2C }}     & 50.2 & 47.3 & 48.8\\
\bottomrule
\end{tabular}}
\caption{Attention analysis.}
\label{tab:attn}
\end{table}


\section{Related Work}
While writing this paper, we noticed that a paper \cite{Du2022LearningTS} that has similar ideas to our work was submitted to the arXiv. They propose a two-stage paradigm for selecting from multiple questions. They also propose a network similar to our {\bf \textsc{\name-B}} to select from multiple options in parallel. We summarize the differences between their work and ours as follows: (1) Different in paradigm. We have an expand stage to further improve the quality of recalled candidates (2) Different in models. {\bf \textsc{\name-S}} and {\bf \textsc{\name-B}} are both different from theirs in both input format and scoring. We additionally propose to discard the C2C attention and study the effect of removing different parts of attention. (3) We focus more on entity typing and conduct extensive experiments covering two languages and three settings (ultra-fine-grained, fine-grained, and coarse-grained). We analyze the effect of using different PLM backbones for a fairer and more comprehensive comparison.
The paradigm of our work is also inspired by works in entity linking and information retrieval. \citet{wu2019zero} uses a retrieval and rerank paradigm for entity linking, they first generate entity candidates using a bi-encoder and rerank them using a vanilla cross-encoder. Our paradigm with an additional expand stage and our proposed {\bf \textsc{\name}} models are also potentially useful for entity linking. We leave it for future work. \citet{halter} represents the query document and candidate documents as vectors and proposed to use a transformer to rerank all candidate documents in parallel for passage retrieval. Compared to them, we tackle entity typing and preserve all information of mention and context rather than represent them as a single vector, the paradigm, model architecture, and training objective are also different.

\section{Conclusion}
In conclusion, we propose a recall-expand-filter paradigm for ultra-fine entity typing. We train a recall model to generate candidates and use MLM and exact match to improve the quality of recalled candidates, then use filter models to obtain final type predictions. We also propose a filter model called multi-candidate cross-encoder ({\bf \textsc{\name}}) to concurrently encode and filter all candidates and study the influences of different input formats and attention mechanisms. Extensive experiments on entity typing show that our paradigm is effective, and the {\bf \textsc{\name}} models under our paradigm reach SOTA performances on both English and Chinese UFET datasets and are also very effective on fine and coarse-grained entity typing.  {\bf \textsc{\name}} models have comparable inference speed to simple ({\bf \textsc{\name}})  models and are thousands of times faster than previous SOTA cross-encoders.

\bibliography{main}
\bibliographystyle{acl_natbib}

\end{document}